\documentclass[a4paper]{article}

\usepackage{INTERSPEECH2019}
\usepackage{amsmath,graphicx}
\usepackage[usenames, dvipsnames]{color}
\usepackage{booktabs}
\usepackage{multirow}
\usepackage{array}
\title{The Marchex 2018 English Conversational \\ 
	Telephone Speech Recognition System}
\name{Seongjun Hahm, Iroro Orife, Shane Walker and Jason Flaks}
\address{
  Marchex Inc., 520 Pike, Seattle, WA 98101, USA}
\email{\{shahm,iorife,spwalker,jflaks\}@marchex.com}

\begin{document}
\maketitle
\begin{abstract}
	In this paper, we describe recent performance improvements to the production Marchex speech recognition system for our spontaneous customer-to-business telephone conversations. In our previous work, we focused on in-domain language and acoustic model training. In this work we employ state-of-the-art semi-supervised lattice-free maximum mutual information (LF-MMI) training process which can supervise over full lattices from unlabeled audio. On Marchex English (ME), a modern evaluation set of conversational North American English, we observed a 3.3\% (3.2\% for agent, 3.6\% for caller) reduction in absolute word error rate (WER) with 3x faster decoding speed over the performance of the 2017 production system. We expect this improvement boost Marchex Call Analytics system performance especially for natural language processing pipeline.
	
\end{abstract}

\section{Introduction}
\label{sec:intro}
Marchex's call and speech analytics business handles over one million calls per one business day, analyzing decades of audio per week. These spontaneous conversational, consumer-to-business phone calls occur on modern mixture of mobile phones and landlines, capturing everyday North American dialog in every possible accent variant, speech rate, English language fluency, speaker demographic, under broad environmental conditions, with a comprehensive, colloquial vocabulary. Although there have been incredible performance improvements on large vocabulary continuous speech recognition (LVCSR) task \cite{xiong2018microsoft,saon2017english,han2017capio}, these variabilities, which vary over time, often obscure the true performance of a production automatic speech recognition (ASR) system due to inconsistencies between the training and test-set data. In real world, test-set data is rapidly changing as new products and businesses come up.

To more fully harness the new data and scale of Marchex call traffic, this paper elaborates on the semi-supervised approach that we introduced in \cite{walker2017semi} to substantially increase the quantity and quality of telephone speech transcripts available to train a modern, production-ready conversational LVCSR system. Furthermore, this approach enables us to maintain the production system performance to be stable from unavoidable time-varying real test-set data.

For acoustic modeling, it is valuable to have such a large and varied dataset which captures diverse language contexts, noise conditions as well as changes in acoustic channel features based on shifts in device and codec technology. To obtain better quality transcripts, we re-decoded the original unsupervised dataset of 30,000 hours of audio, using the production \emph{online-nnet2} model. Post-processing and re-selecting the most suitable utterances yielded a new 5,500 hour dataset \cite{walker2017semi} that was used to train a series of time-delay neural network (TDNN) acoustic models with the LF-MMI sequence objective function \cite{povey2016purely}. This data is decoded transcription chosen by confidence measure and perplexity threshold. Because this is not manual transcription by human, we cannot guarantee whether the 1-best transcription is the correct or not. This could be critical issue for discriminative training which mainly depends on reference transcription. In this sense, we expect that full lattice-based semi-supervised LF-MMI training uses these data more appropriately than 1-best approach. In this work, we fixed 5,500 hour dataset as training data to verify performance difference only from acoustic model difference.


Our objectives in this work is to obtain the lowest possible word error rate (WER) production system on the ME evaluation dataset. Unlike other efforts \cite{saon2015ibm,saon2016english,xiong2018microsoft,saon2017english,han2017capio} that explicitly eschew practical considerations such as speed (decoding real-time factor) and memory consumption (model size), we have concentrated on making the best use of our data scale to efficiently improve and maintain our production ASR accuracy. The importance of ASR accuracy with respect to downstream natural language processing (NLP) model performance can be understood in two ways. 1) Some types of mis-transcriptions characterized by admissible substitutions can be learned by downstream classifiers. To the extent that the sequence is not distinct from other conversational utterances, precision will suffer. 2) Mis-transcriptions characterized by deletions, especially where monosyllabic words are key features, lead to downstream models that learn spurious relationships \cite{kumar2014some}. 

The paper is organized as follows, Section \ref{sec:system-improvements} will provide an overview of the speech recognition system, detailing the updates to the acoustic model training regime. Section \ref{sec:experiments} will summarize the experiments and results while Section \ref{sec:discussion} provides perspective on lessons learned and directions for future work.

\section{System Improvements}
\label{sec:system-improvements}

Marchex call processing servers receive two channels of 8kHz $\mu$-law encoded audio for the caller and agent \emph{call-legs}. Both channels are individually segmented by an online Voice Activity Detector (VAD), creating single utterances that are stored in a large Apache Kafka \emph{audio topic}, each transcribed in turn by a fleet of Amazon Web Services (AWS)-based production Kaldi \cite{povey2011kaldi} hosts. Each Kaldi hosts runs a decoder using a deep neural network acoustic model and utilizes a decoding graph with an enhanced lexicon and a very large, conversational language model (LM) \cite{walker2017semi}.

Figure 1 shows the sequence of training operations. The seed model is ``pre-built'' deep neural network - hidden Markov model (DNN-HMM) hybrid model provided with Kaldi\footnote{This seed model can be downloaded from http://kaldi-asr.org/downloads/build/8/trunk/egs/fisher\_english/.}. In preparation to improve the existing \emph{online-nnet2} training regime, the lexicon was enhanced with pronunciation probabilities for words with multiple phonetic pronunciations and silence probabilities between phonemes within a word were also added \cite{peddinti2015time}. Next, acoustic model training recipes were updated to use Kaldi \emph{nnet3} TDNN architectures \cite{povey2016purely,manohar2018semi,povey2018semi}. 

The tools and training recipe for our experiments are based on the semi-supervised LF-MMI work by Manohar et al. \cite{manohar2018semi} who use the Fisher English \cite{cieri2004fisher} dataset with a seed model trained on purely supervised (only hand-labeled) transcripts. By contrast, our TDNN seed model was trained on semi-supervised data (5,500 hours). The supervised data is only 14 hours among 5,500 hours of training data. Because our previous model is also trained using semi-supervised approach with 1-best decoded result (transcript), we expect to reduce WER considerably.


\subsection{Semi-supervised acoustic modeling}
\label{sec:acoustic-modeling}

A TDNN is a kind of feed-forward neural network architecture shown to be effective in handling long range temporal dependencies. Each layer operates at a different temporal resolution with initial layers processing smaller contexts, while deeper layers attend to wider temporal contexts. TDNNs improve on limitations of traditional recurrent neural networks (RNNs) i.e.\ high computational complexity that is non-parallelizable, with a sub-sampling technique where each layer's input is selected from specific time steps in the previous layers. By carefully selecting splicing indices, per-layer computation is reduced while assuring an adequate amount of temporal context is seen by each layer of the network \cite{peddinti2015time, povey2016purely}.

In moving to new \emph{nnet3} TDNN recipes, a number of updates were made along the way. These included overhauling the non-linearity function from p-norm to ReLU, applying dropout and tuning the size of the left/right context and the dimension of the output layer. The final recipes also utilized a factorized TDNN or TDNN from \cite{povey2018semi}, which improves on the state of the art TDNN+LSTM (long short term memory), by constraining one of each weight matrix's two factors to be semi-orthogonal. TDNN-F also employs a ResNet-style skip connection to concatenate the output of previous non-adjacent layers.

\begin{figure}[t]
	\begin{center}
		\includegraphics[width=3.2in]{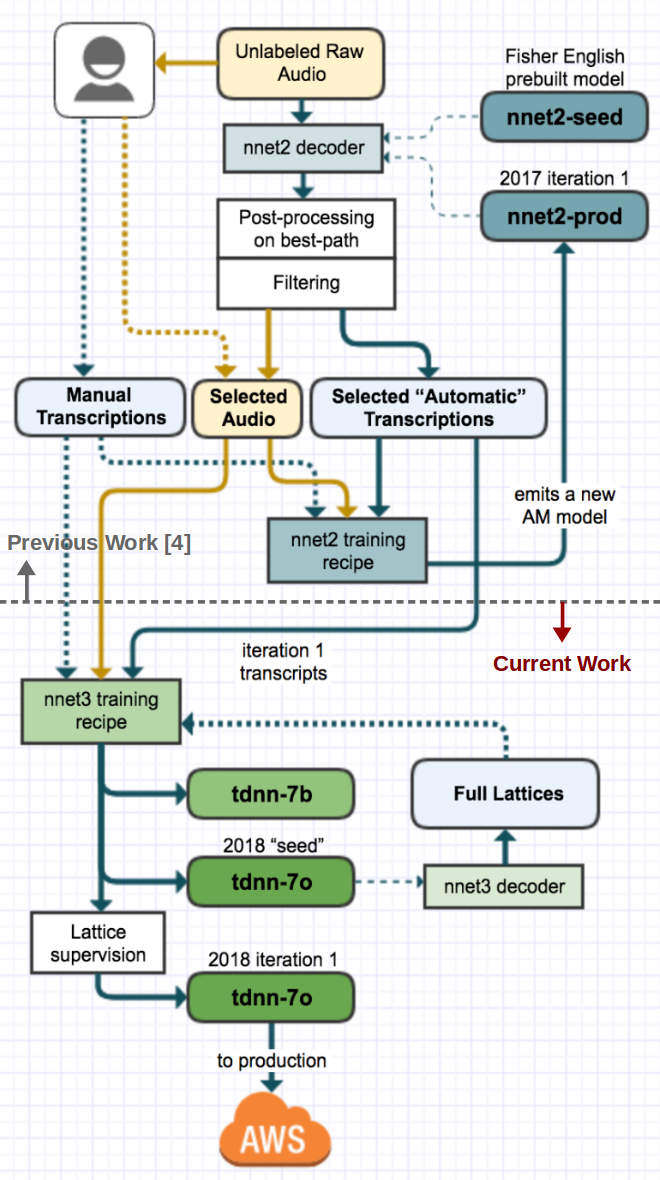}
		\caption{AM training process}
	\end{center}
	\vspace{-7mm}
\end{figure}

\subsubsection{LF-MMI Training}
We will give a brief overview of LF-MMI training with semi-supervised lattice supervision. For a full description, the reader is assigned an exhaustive study of \cite{povey2011kaldi, peddinti2015time, povey2016purely}. For acoustic modeling, maximum mutual information (MMI) is an objective function used in sequence discriminative training of neural networks. Word lattices are generated from the training data using a frame-level, cross-entropy pre-trained Gaussian mixture model (GMM) model and a ``weak" language model. There are two sets of lattices, a numerator and denominator lattice. The numerator represents the ``alignment" of the correct sentence, while the denominator lattice embodies all possible word sequences from the recognizer and is generated using a very big beam. The denominator lattice in the training objective is the same for all utterances. Optimization involves gathering statistics and re-estimating model parameters using the forward-backward algorithm over these word-level lattices. 

In LF-MMI, sequence discriminative training is done from ``scratch", at the finer \emph{phone} resolution, without using word-lattices. A GMM or deep neural network-hidden Markov model (DNN-HMM) system may be used generate lattices from which phone graphs are derived. For the MMI objective, there are now numerator and denominator graphs which are stored as Finite State Acceptors (FSA). Accordingly, there still needs to be a summation over all possible label sequences, so a 4-gram \emph{phone-level} LM, in lieu of a word-level LM, is used to create the denominator graph, which is post-processed to be as small as possible for practical on-GPU training.

Finally, within a LF-MMI training regime with lattice-based supervision, a seed LF-MMI decoder generates decoding lattices from unlabeled audio. These lattices are rescored with a strong LM, incorporated with tolerances, converted to phone-level graphs then per-utterance FSTs. Finally they are split into numerator FSTs of fixed lengths for use in LF-MMI training. In this case tolerances represent the ``slack" or ``wiggle room" given to each phone's alignment in a window around where it was defined in the lattice.

The intuition is that these graphs, from unsupervised audio, contain alternative paths, expressing a level of uncertainty that is useful to propagate back into the model training and optimization phase. By not collapsing these alternatives, the optimization and strong LM rescoring steps have a better opportunity to more accurately supervise.

\begin{table}[!ht]
	\caption{A summary of training condition} \label{tab:model-summary}
	\vspace{-3mm}
	\begin{center} 
		\begin{tabular}{|l|c|c|} \hline
			&  \multicolumn{1}{c|}{Subset} &  \multicolumn{1}{c|}{Full-set} \\ \hline
			Sample Rate &  \multicolumn{2}{c|}{8kHz}   \\\hline
			Feature vector & \multicolumn{2}{c|}{40d MFCC + 100d iVector} \\ \hline
			Frame length & \multicolumn{2}{c|}{25 ms} \\ \hline
			Frame hopsize & \multicolumn{2}{c|}{10 ms} \\ \hline
			\footnotesize {Cepstral Mean Normalization} & \multicolumn{2}{c|}{No} \\ \hline
			Language Model & \multicolumn{2}{c|}{3-gram (KN smoothing)} \\ \hline
			Lexicon & \multicolumn{2}{c|}{134,720 words} \\ \hline
			Total \# utterances & 883,696 & 11,724,400 \\ \hline
			Total speech audio hours & 548.1 & 5,526.5 \\ \hline
			Total \# of speakers & 346,282 & 507,461 \\ \hline
			Input Layer Dimension & \multicolumn{2}{c|}{140} \\ \hline
			\# of Hidden Layers & \multicolumn{2}{c|}{11} \\ \hline
			Output Layer Dimension\footnotemark & 4,856 & 6,358 \\ \hline
		\end{tabular}	
	\end{center}
	\vspace{-5mm}
\end{table} 
\footnotetext{There are the same number of output layer dimension across the different type of models because the models share tree structure.}

\section{Experiments}
\label{sec:experiments}
To verify effectiveness of LF-MMI lattice-supervision, we performed semi-supervised LF-MMI experiments using 5,500 hours training data. All experiments were performed using Kaldi toolkit \cite{povey2011kaldi}.

\subsection{Experiments on a 550 hour subset}
To validate suitability of LF-MMI lattice-supervision, we performed preliminary experiments before we build the final production model. Table \ref{tab:model-summary} shows a summary of training condition. Utterances were chosen based on the criteria that there be sufficient leading and trailing time on each speech segment, i.e. are correctly segmented. The model was trained with 14 supervised hours and 534 unsupervised hours of data. iVector extractor was also trained on the combined datasets. Our testset is chosen from ME, a modern-day North American English conversational task comprised of 7,000 utterances or 4.5 hours of no-filler, manually transcribed conversational audio, sourced from more than 3,000 calls. 

\begin{table*}[!ht]
	\vspace{-2mm}
	\caption{Preliminary experimental results using 550 hour subset (1-best semi-supervised). LM is the same as across the models.} \label{tab:500h}
	\vspace{-3mm}
	\begin{center}
		\begin{tabular}{|c|c|c|r|r|r|r|} \hline 
			\textbf{AM} & Test set & WER \% (3-gram LM) & \small \# Model Params & AM Size & WFST Size & Decoding RTF \\ \hline
			\multirow{3}{25mm}{\centering\textbf{TDNN}\\\texttt{tdnn\_7b}} & Agent& 10.4 &  & & & \\ \cline{2-3} 
			& Caller & 14.3 & 11.4 M & 44 MB & 370 MB & 0.29\\ \cline{2-3}
			& Combined & 12.1 & &  & & \\ \hline 
			
			\multirow{3}{25mm}{\centering\textbf{TDNN+LSTM}\\\texttt{tdnn\_lstm\_1b}}& Agent & 10.7 & &  & & \\ \cline{2-3}
			& Caller & 14.0 & 18.6 M & 72 MB & 370 MB & 0.57 \\ \cline{2-3}
			& Combined & 12.2 & &  & & \\ \hline 
			
			\multirow{3}{25mm}{\centering\textbf{bGRU}\\\texttt{bgru\_1a}} & Agent & \hspace{2mm}\textbf{9.3}  & &  & &  \\ \cline{2-3}
			& Caller & \textbf{12.6} &  47.5 M & 183 MB & 370 MB & 2.39 \\ \cline{2-3}
			& Combined & \textbf{10.7} & & & &  \\ \hline 
			
			\multirow{3}{25mm}{\centering\textbf{TDNN}\\\texttt{tdnn\_7o}} & Agent & \hspace{2mm}9.9 & & & &   \\ \cline{2-3}
			& Caller & 13.0 &  22.2 M & 86 MB & 370 MB & 0.24 \\ \cline{2-3}
			& Combined  & 11.3  & & &  & \\ \hline 
			
%
		\end{tabular}	
	\end{center}
	\vspace{-3mm}
\end{table*}

\begin{table*}[!ht]
	\caption{Preliminary experimental results using 550 hour subset (1-best and full lattice-based approaches). LM is the same as across the models.} \label{tab:500h-semi}
	\vspace{-3mm}
	\begin{center}
		\begin{tabular}{|c|c|c|r|r|r|r|} \hline 
			\textbf{AM} & Test set & WER \% (3-gram LM) & \small \# Model Params & AM Size & WFST Size & Decoding RTF \\ \hline
			\multirow{3}{25mm}{\centering\textbf{TDNN}\\\texttt{tdnn\_7o}} & Agent & \hspace{2mm}\textbf{9.3} & & & &   \\ \cline{2-3} 
			& Caller & 12.3 &  22.2 M & 86 MB & 370 MB & 0.24 \\ \cline{2-3}
			& Combined & 10.6  & & &  & \\ \hline
			
			\multirow{3}{25mm}{\centering\textbf{TDNN}\\\texttt{tdnn\_7o}\\(full lattice)} & Agent & \hspace{2mm}\textbf{9.3} & & & &  \\ \cline{2-3}
			& Caller & \textbf{11.9} & 22.2 M & 86 MB & 370 MB & 0.26 \\ \cline{2-3}
			& Combined & \textbf{10.4} & & & & \\ \hline
		\end{tabular}	
	\end{center}
   \vspace{-3mm}
\end{table*}

We trained 4 different kinds of models, TDNN  (\texttt{tdnn\_7b}), TDNN-LSTM (\texttt{tdnn\_lstm\_1b}), bi-directional gated recurrent unit (bGRU; \texttt{bgru\_1a\footnote{We did make small change (putting right side GRU-layer instead of TDNN-layer) based on \texttt{tdnn\_opgru\_1b}. }}) and factorized TDNN-F (\texttt{tdnn\_7o}). Table \ref{tab:500h} shows the results to select the model structure for the production. We considered WER, the number of model parameters, the size of acoustic model (AM), the size of weighted finite state transducer (WFST), and decoding speed (real time factor; RTF) to select the model for semi-supervised experiment. We want to note that this experiment is also semi-supervised training. The reference transcriptions for unsupervised data are from the decoded results by using our production \emph{online-nnet2} model. RTF which is slower than real-time (2.39 for bGRU in Table 2) is not appropriate for the production model even the model shows the best performance. Therefore we selected TDNN (\texttt{tdnn\_7o}) based on the results (WER: 2nd, the number of model parameters: 2nd, AM size: the 3rd, WFST size: the same across the models, RTF: 1st). Then using TDNN (\texttt{tdnn\_7o}) as the model for re-decoding, we performed semi-supervised experiment with 1-best and full-lattice based approach. 

Table \ref{tab:500h-semi} shows the results for comparing 1-best and full lattice-based semi-supervised approaches. Here the WER gain TDNN model (\texttt{tdnn\_7o}) from the same model in Table 2 is from better quality transcription (one more semi-supervised training with the same training data). Even though there is no performance difference on agent side between two models, we observed 0.4\% absolute (3.3\% relative) WER improvement on caller side. We could think this is small improvement. However, in real production system, this could be huge difference if we consider over one million calls per one business day. Furthermore, this can be a meaningful difference for NLP modeling of Marchex Call Analytics system.

The interesting results we found was full-lattice based semi-supervised TDNN model (\texttt{tdnn\_7o}) using 550 hours of training data outperformed our production \emph{online-nnet2} model, DNN (\texttt{nnet2}) in Table 3, which is trained using 5,500 hours of training data. The actual gains are 1.0\% for agent channel and 1.3\% for caller channel, respectively. We expect more gain using full 5,500 hours of training data. We repeat these experiments using full 5,500 hours training data.



\subsection{Experimental results}
\label{sec:experimental results}
Table \ref{tab:results} shows the results of semi-supervised acoustic modeling described in Section \ref{sec:acoustic-modeling}. The testset is the same as 550 hour subset experiment (4.5 hours). We report WER figures, per call-channel as well results from rescoring with a 4-gram LM and a Tensorflow LSTM LM (TF-LSTM). 

\begin{table*}[!ht]
	\begin{center}
		\vspace{1mm}
		\caption{WER performance and model attributes for 5,500 hours full training set. LM is the same as across the models.} \label{tab:results}
		\vspace{-1mm}
		\begin{tabular}{|c|c|r|r|r|r|r|r|r|} \hline
			\multirow{3}{10mm}{\centering \textbf{AM}}& \multirow{3}{10mm}{\centering Test set} & \multicolumn{3}{c|}{\textbf{WER \%}} & \multirow{3}{20mm}{\centering\# Model Params} & \multirow{3}{10mm}{\centering AM Size} & \multirow{3}{15mm}{\centering WFST Size} & \multirow{3}{15mm}{\centering Decoding RTF} \\ \cline{3-5} 
			&&\multicolumn{3}{c|}{LM}&&& & \\ \cline{3-5}
			& & 3-gram &  4-gram & TF-LSTM & & & & \\ \hline
			\multirow{3}{20mm}{\centering\textbf{DNN}\\\texttt{nnet2}} & Agent & 10.3 & 9.8  & 9.8 & & & & \\ \cline{2-5} 
			& Caller & 13.2 & 12.8 & 12.5 & 13.7 M & 54 MB & 1.1 GB & 0.99 \\ \cline{2-5}
			& Combined & 11.5 & 11.1 & 11.0  & & & & \\ \hline
			
			
			\multirow{3}{20mm}{\centering\textbf{TDNN}\\\texttt{tdnn\_7o}} & Agent & 8.3 & 8.1  & 7.8  & & & & \\ \cline{2-5} 
			& Caller & 10.9 & 10.7 & 10.2 & 23.4 M & 91 MB & 370 MB & 0.34 \\ \cline{2-5}
			& Combined & 9.4 & 9.3 & 8.9 & & & & \\ \hline
			
			\multirow{3}{20mm}{\centering\textbf{TDNN}\\\texttt{tdnn\_7o}\\(full lattice)} & Agent & 7.1 & 7.1  & \textbf{7.0} & & & &  \\ \cline{2-5} 
			& Caller & 9.6 & 9.7 & \textbf{9.2} & 23.0 M & 89 MB & 370 MB & 0.30 \\ \cline{2-5}
			& Combined & 8.2 & 8.2 & \textbf{8.0}  & & & &\\ \hline
		\end{tabular}	
	\end{center}
\vspace{-3mm}
\end{table*}

We followed standard scoring method similar to other researches \cite{xiong2018microsoft,saon2017english,han2017deep}. There were a scoring changes from our previous results reported in \cite{walker2017semi}. Those are partial words \{gue- vs. guess, guest\}, multiple words \{firestone vs. fire stone\}, colloquial forms \{going to vs. gonna\}. Hesitations and filler words \{uh, um, ah, er\} were also removed. In Table 4, you can see the baseline performance is 10.3\% and 13.2\%, agent and caller respectively, based on new scoring (14.3\% and 17.5\%, agent and caller respectively in \cite{walker2017semi}). 

The seed for each model is the model in the previous line. The initial seed model for DNN (\texttt{nnet2}) is the DNN (\texttt{nnet2}) model provided with Kaldi (see footnote 1).

In Table 4, 4-gram results are better than 3-gram LM. We also verified TF-LSTM LM outperformed 4-gram LM. However, we could not observed noticeable performance improvement on full lattice-based semi-supervised TDNN (\texttt{tdnn\_7o}). TF-LSTM LM rescoring takes time and we need to put additional neural network model to production. So we decided 1st-pass decoding full lattice-based semi-supervised TDNN (\texttt{tdnn\_7o}) as production model.

We compare the result with 3-gram LM here. The semi-supervised (1-best) TDNN (\texttt{tdnn\_7o}) showed 2.0\% (agent), 2.3\% (caller) absolute WER improvement over DNN (\texttt{nnet2}) model. This is significant improvement exceeding our expectation. This gain comes mostly from semi-supervised LF-MMI approach. For the semi-supervised (full-lattice) TDNN (\texttt{tdnn\_7o}) showed absolute WER improvements 1.2\% and 1.3\% absolute WER improvement for agent and caller channel, respectively over the semi-supervised (1-best) TDNN (\texttt{tdnn\_7o}). This shows how factorized TDNN significantly improved quality of transcription data for training. And this also shows that the full lattice-based approach is effective especially for a semi-supervised training process. Finally we observed a 3.3\% (3.2\% for agent, 3.6\% for caller) reduction in absolute word error rate (WER) with 3x faster decoding speed over the performance of the 2017 production system, DNN (\texttt{nnet2}). 

Based on the overall experimental results, we could find the fact that WER of agent channel was better than WER of caller channel. This is obvious because our LM contains some portion of agent sentences frequently spoken whereas caller channel (from customer side) sentences are usually hard to predict. Another reason is that quality of agent channel audio is usually better than caller channel which can usually contain background noise (e.g., babble, street, car, etc).

From Table \ref{tab:results}, we capture the improvements to the decoding real-time factor and graph size, which translates into a significantly smaller production decoding fleet size and lower CPU as considering our daily call volume.

\section{Discussion}
\label{sec:discussion}
In this report we have chronicled improvements made to our production ASR system based on a semi-supervised LF-MMI training with lattice-based supervision. Our current directions involve experiments to dynamically add words to the FST graph using word-class language models \cite{horndasch2016combining,horndasch2016how}. Other practical solutions involve creating customer and vertical specific training and evaluation sets, including customer-specific LM-rescoring. 

While there is a lot of hype and promise around end-to-end (E2E) systems like ESPNET \cite{kim2017joint, watanabe2017hybrid} and DeepSpeech \cite{amodei2015deep}, our practical experience experimenting and training these models has shown them to be very sensitive to the data cleanliness in ways that encumbers a semi-supervised training process. Furthermore, in comparison with current LF-MMI techniques, E2E performance is not ready for production for the spontaneous and large vocabulary conversational task.

Finally, since the original 30,000 hours of unlabeled audio originally used in \cite{walker2017semi}, we have grown the size of our unsupervised audio dataset to be over 80,000 hours. Future experiments involve using this larger, unfiltered dataset as the starting point for semi-supervised model training efforts.

\section{Acknowledgement}
\label{sec:acknowledgement}

The authors thank the following former Marchex colleagues for their efforts and contributions to the Marchex Speech Recognition System: Morten Pedersen and Hari Rajagopal.

\bibliographystyle{IEEEtran}



\end{document}